\documentclass[letterpaper, 10 pt, conference]{ieeeconf}  

\IEEEoverridecommandlockouts                              

\usepackage{graphics} 
\usepackage{epsfig} 
\usepackage{amsmath} 
\usepackage{makecell}
\usepackage{verbatim}
\usepackage{amsmath}

\usepackage[disable]{todonotes}

\usepackage[normalem]{ulem} 
\usepackage{subfigure}

\newcommand{\wenzhen}[1]{\todo[inline,color=red!40]{Wenzhen: #1}}
\newcommand{\joe}[1]{\todo[inline,color=yellow!40]{Joe: #1}}
\newcommand{\feng}[1]{\todo[inline,color=blue!40]{Xiaofeng: #1}}

\title{\LARGE \bf Estimating Properties of Solid Particles Inside Container \\Using Touch Sensing 
}

\author{Xiaofeng Guo$^{1}$, Hung-Jui Huang$^{1}$, and Wenzhen Yuan$^{1}$
\thanks{This work is supported by the Toyota Research Institute (TRI).}
\thanks{$^{1}$Xiaofeng Guo, Hung-Jui Huang, and Wenzhen Yuan are with the Robotics Institute, Carnegie Mellon University, 5000 Forbes Ave, Pittsburgh, PA 15213, USA {\tt\small\{xguo2, hungjuih, wenzheny\}@andrew.cmu.edu}}%
}

\begin{document}

\maketitle
\thispagestyle{empty}
\pagestyle{empty}

\begin{abstract}
Solid particles, such as rice and coffee beans, are commonly stored in containers and are ubiquitous in our daily lives. Understanding those particles’ properties could help us make later decisions or perform later manipulation tasks such as pouring. Humans typically interact with the containers to get an understanding of the particles inside them, but it is still a challenge for robots to achieve that. This work utilizes tactile sensing to estimate multiple properties of solid particles enclosed in the container, specifically, content mass, content volume, particle size, and particle shape. We design a sequence of robot actions to interact with the container. Based on physical understanding, we extract static force/torque value from the F/T sensor, vibration-related features and topple-related features from the newly designed high-speed GelSight tactile sensor to estimate those four particle properties. We test our method on $37$ very different daily particles, including powder, rice, beans, tablets, etc. Experiments show that our approach is able to estimate content mass with an error of $1.8$ g, content volume with an error of $6.1$ ml, particle size with an error of $1.1$ mm, and achieves an accuracy of $75.6$\% for particle shape estimation. In addition, our method can generalize to unseen particles with unknown volumes. By estimating these particle properties, our method can help robots to better perceive the granular media and help with different manipulation tasks in daily life and industry.
\end{abstract}

\section{INTRODUCTION}
\label{Sec: intro}



Granular media, also known as solid particles, are widely used in daily life, from construction materials like sand and stones, to culinary ingredients like spices powder and beans. Understanding their properties is important for humans to better utilize them. However, particles are commonly stored in containers like jars or bottles in daily life, making them challenging to perceive. Instead of pouring particles out and using vision, humans often use touch sensing to tell what is inside the container by interacting with the container dynamically. For example, by shaking the container, humans could roughly tell whether the particle is large or small; whether the particle is ball-shaped or irregular-shaped. However, it remains challenging for robots to estimate the property of particles in that case. 

In this work, we target estimating the property of the solid particles inside a container. Specifically, we try to estimate the content mass, content volume, particle size, and particle shape. We choose these four properties because they describe the basic geometry information of universal solid particles. The estimation of these properties helps people recognize the particles, whether they are seen or unseen, and helps to complete the following manipulation tasks such as pouring.

Previous efforts attempt to classify different solid particles enclosed in the container using audio \cite{eppe2018deep}, tactile \cite{chen2016learning}, reflectance spectroscopy \cite{hanson2022slurp}, or multi-modalities \cite{sinapov2014learning}. They manually designed the actions to be rotation \cite{jin2019open}, shaking \cite{chen2016learning}, squeezing \cite{guler2014s}, and so on. Although most of them achieved a high classification accuracy, they have poor generalizability and are hard to handle unseen particles. Most of them also require the volume of the particle to be the same as during training. To the best of our knowledge, there are no previous works on the estimation of the properties of the solid particles enclosed in the container, especially particle size and shape. In this work, we aim to estimate some basic particle properties so the unseen particles can be better described, which is a challenging task due to the complex dynamic of solid particles. Multiple particle properties such as density, humidity, and texture jointly influence the particle macro behavior, of which a precise model is still missing. Therefore, finding a perceptible particle behavior which directly related to individual particle property remains challenging.

\begin{figure}
    \centering
    \includegraphics[width = 0.98\linewidth]{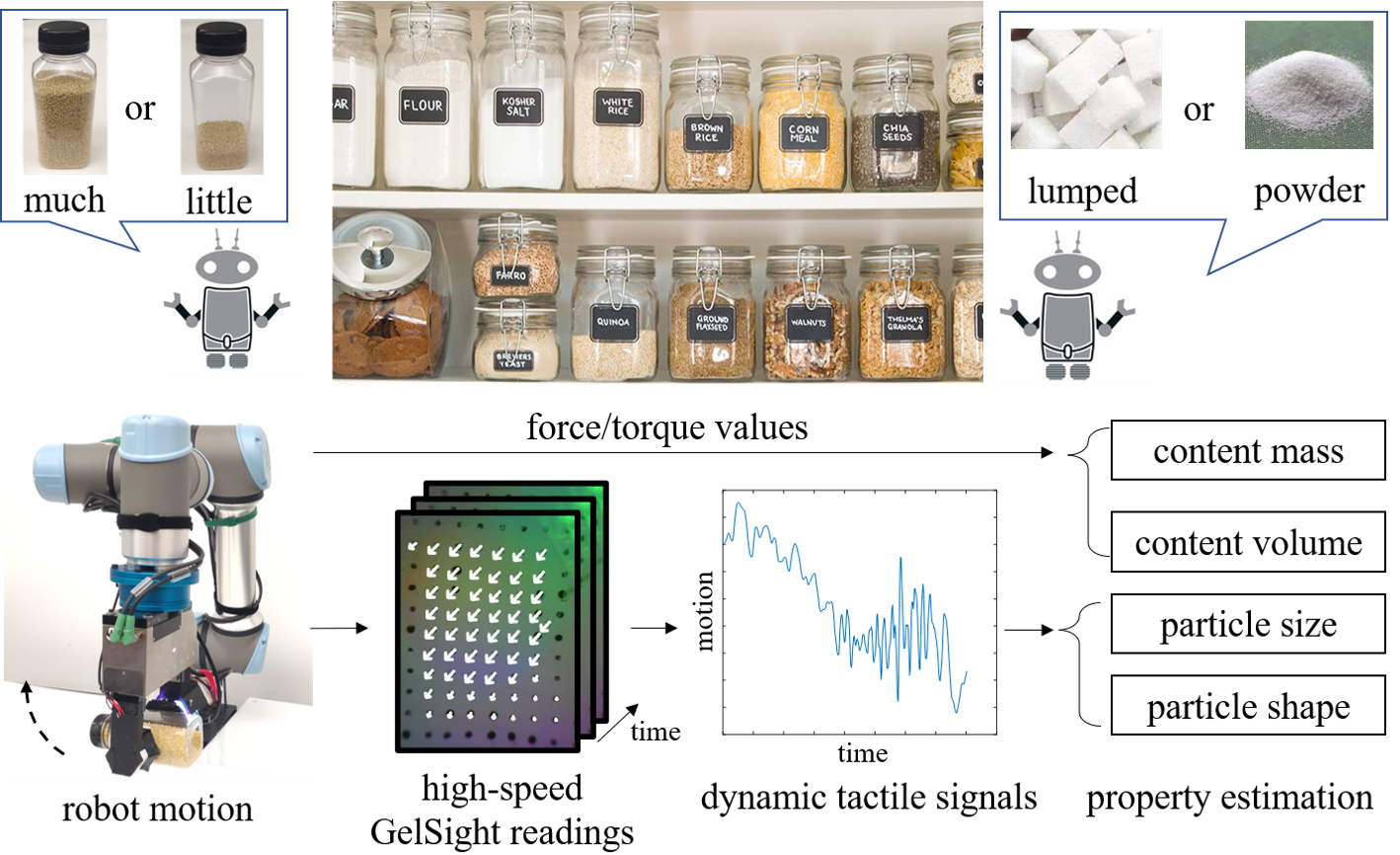}
    \caption{We use tactile sensing to estimate the properties of particles in a container, including content mass, content volume, particle size, and particle shape. We use different actions to interact with the container and collect the tactile data from both the F/T sensor and our newly designed high-speed GelSight tactile sensor.}
    \label{fig:concept}
\end{figure}

In this work, we propose a pipeline to use tactile sensing to estimate different properties of solid particles enclosed in the container, including content mass, content volume, particle size, and particle shape, as shown in Fig. \ref{fig:concept}. We utilize both static tactile sensing, which focuses on the F/T at a static state, and dynamic tactile sensing, which focuses on the temporal changing of tactile signals. To achieve better fingertip dynamic tactile sensing, we design a new tactile sensor: high-speed GelSight, which has both high spatial resolution ($640\times480$) and high temporal resolution ($815$ Hz). To acquire tactile data, we design a sequence of robot motions to interact with the container. The robot first lifts and tilts the container to different angles. The content mass and content volume are first estimated by measuring the static wrist F/T values at each rotation angle. Then the robot rotates the container in two directions at different speeds. We extract the vibration-related features and topple-related features separately from the dynamic tactile signals recorded from the new-designed high-speed GelSight during these two robot motions. We utilize these features to estimate the particle size and shape.

The main contribution of this work is proposing a framework to solve a newly defined task: property estimation of the solid particles inside a container. To describe the particles inside the container, we estimate the content mass, content volume, particle size, and particle shape. We also contribute to designing a new tactile sensor: high-speed GelSight, which has both high temporal resolution and high spatial resolution. We collected the data from $37$ very different daily particles with different amounts in the container. The experiments show that our approach can estimate content mass with an error of $1.8$ g, content volume with an error of $6.1$ ml, particle size with an error of $1.1$ mm, and particle shape with an estimation accuracy of $75.6$\%. We also show that our method can be generalized to unseen daily particles with unseen volume. We believe our method can improve the ability of robots to perceive solid particles and help them perform various manipulation tasks in daily life and industry.

\wenzhen{I think the summarization of the contributions looks weak. E.g., I'm not convinced that the task definition in this specific case looks like a significant contribution. Besides, it's the first time you mention the high-speed GelSight. You probably want to highlight it earlier. You want to highlight what's the major achievement you make through this work.} \joe{agreed}




\section{RELATED WORKS}
\label{sec: relatedworks}

\subsection{Properties of Granular Media}

The study of granular media properties is an active research area within the physics community. Some work concluded the relationship between the granular media macroscale behavior to particle-scale properties, which inspires us when we define the robot motion. The angle of repose (AoR) is one common metric to describe particle behavior. It is commonly defined to be the steepest slope of the unconfined material, measured from the horizontal plane on which the material can be heaped without collapsing \cite{mehta1994dynamics}. AoR was found to be related to multiple particle properties of granular material, including the particle shape \cite{dai2016effect}, the particle size \cite{zhou2002experimental}, \cite{botz2003effects}, the particle friction coefficients \cite{santos2016investigation}, the content moisture \cite{zaalouk2009effect}, the number of particles \cite{miura1997method}, et al. Although multiple works concluded the relationship between AoR and particle properties, few works applied it to robot applications. In our work, we design one robot motion inspired by this and extract topple-related features which reflect the particles' AoR to estimate solid particle properties.


Very few previous works solved the problem that using robots to estimate solid particle properties. The only relevant work we found was by Matt \cite{matl2020inferring}, who used the pattern of the particle stacked rings to calibrate parameters of the particles, such as friction and restitution, for simulation. In our work, we define the particle property estimation task from a different perspective: we focus on the estimation of some explicit particle properties such as particle size and particle shape. These properties are more intuitive for humans and robots to get an understanding of those particles.

\subsection{Content Classification}

Although very few works targeted estimating particle properties, researchers have tried to recognize the particle type in opaque containers, which is a relevant easier task. Those methods provide us with some inspiration about how to distinguish particles with different properties. Those works used dynamic signals to recognize the content, which are collected with the robot motion such as shaking, dropping, and tapping \cite{sinapov2009interactive}, \cite{sinapov2011interactive}, \cite{griffith2012object}. The signals come from either audio or touch. For example, Eppe \cite{eppe2018deep} presented a strategy where the robot shakes the plastic capsules and uses auditory information to classify the contents. Jin \cite{jin2019open} collected the sound generated by rotating the bottles with different particles to classify the contents. Clarke \cite{clarke2018learning} used shaking sounds for estimating flow and amounts. Sinapov \cite{sinapov2014learning} proposed a framework to detect object categories by combining visual, auditory, and proprioceptive information. They explored 36 objects with different contents, weights, and colors by 10 behaviors. Chen \cite{chen2016learning} combined acoustic and acceleration information for content classification. There are also some works trying to estimate liquid properties and dynamic tactile sensing is found to be useful. Huang \cite{Huang-RSS-22} used dynamic tactile sensing to estimate liquid viscosity and height. Saal \cite{saal2010active} used tactile sensing to predict the liquid viscosity and optimized the shaking frequency and the rotation angle of shaking. Matl \cite{matl2019haptic} used F/T data and a model-based method for liquid property estimation and utilized the estimation for accurate pouring liquids.  Different from the above works, we target a more challenging task: estimating the quantity results of solid particle properties. Compared to semantic labels, the quantitative particle property estimation would be more helpful for following manipulation tasks and need to be generalized to unseen particles.

\section{PROBLEM STATEMENT}
\label{sec: problemstatement}
\begin{figure}
    \centering
    \includegraphics[width = 0.95\linewidth]{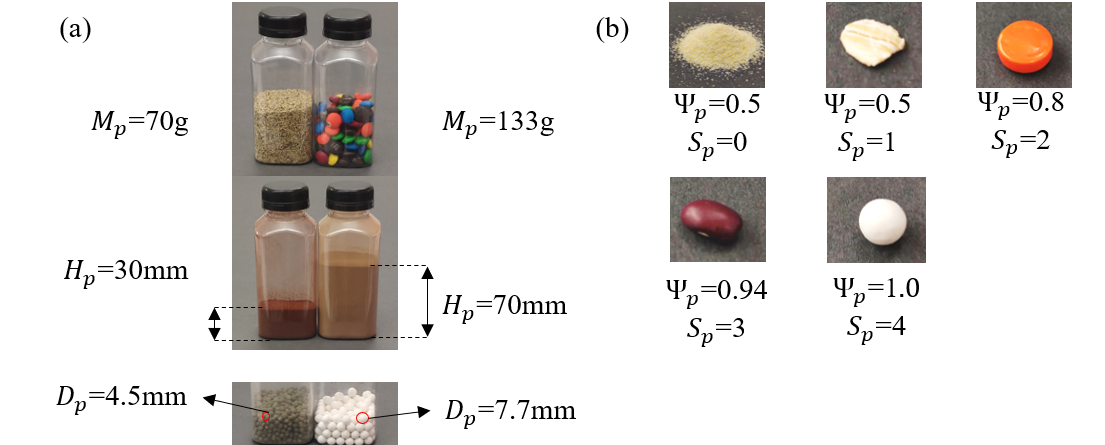}
    \caption{Particle properties to be estimated. (a) Examples of contents with different content mass $M_p$, content volume $V_p$, and particle size $D_p$. $V_p$ is represented as the particle stacking height $H_p$. (b) Examples of particles with different sphericity $\Psi_p$ and the shape descriptor $S_p$. }
    \label{fig:problem_statement}
\end{figure}

\wenzhen{You need to discuss whether the method can be generalized to other containers}

We aim to estimate the properties of solid particles in the container using only touch sensing. In this section, we define the targeting particle properties as the following: content mass $M_p$, content volume $V_p$, particle size $D_p$, and particle shape $S_p$. Some samples of the particles with different properties are shown in Fig. \ref{fig:problem_statement}. These properties are chosen because they are fundamental information to define the particles inside the container and useful perception information for manipulation tasks. For example, accurate measurement of the total volume of particles is essential for precise volume control in robotic pouring applications.


\noindent \textbf{Content mass $M_p$} is defined as the total mass of the particles inside the container.


\noindent \textbf{Content volume $V_p$} is defined as the total volume of those particles and the room between their stacking. In the following experiment, we use the same container during training and testing. Therefore, $V_p$ is linearly correlated to the height $H_p$ of the particle stacks when holding the container vertically. To simplify, we will use content height $H_p$ to represent $V_p$. 

\noindent \textbf{Particle size $D_p$} is defined as the diameter of the sphere with the same volume as the given particle: $D_p = (\frac{6V}{\pi})^{\frac{1}{3}}$, where $V$ is the volume of the individual particle. 

\noindent \textbf{Particle shape $S_p$} is defined by the particle sphericity $\Psi_p$, which is commonly used to describe how similar the particle shape is compared to spheres. Sphericity is defined as the ratio of the surface area of a sphere with the same volume as the given particle to the surface area of the particle \cite{wadell1935volume}: $\Psi_p = \frac{\pi^{\frac{1}{3}}(6V)^{\frac{2}{3}}}{A}$, where $V$ is the volume of the individual particle and $A$ is its surface area. Sphericity is always a number between $0$ and $1$. The more spherical the shape, the closer its sphericity is to $1$. Since the sphericity distribution of common particles is skewed toward $1$ and is unbalanced, and the sphericity of powder is less meaningful, we re-define the shape descriptor to five classes using the following rules:

\begin{equation*}
S_p = \left\{
\begin{aligned}
0 & \hspace{0.5 in} (D_p \leq 1\text{mm})\\
1 & \hspace{0.5 in} (\Psi_p \leq 0.7 \hspace{0.1in} \text{and} \hspace{0.1in} D_p > 1\text{mm})\\
2 & \hspace{0.5 in} (0.7 < \Psi_p \leq 0.9  \hspace{0.1in} \text{and} \hspace{0.1in} D_p > 1\text{mm}) \\
3 & \hspace{0.5 in} (0.9 < \Psi_p \leq 0.96 \hspace{0.1in} \text{and} \hspace{0.1in} D_p > 1\text{mm}) \\
4 & \hspace{0.5 in} (0.96 < \Psi_p \leq 1 \hspace{0.1in} \text{and} \hspace{0.1in} D_p > 1\text{mm}) \end{aligned}
\right.
\end{equation*}
 



\section{METHOD}
\label{sec: method}

To estimate the content mass, volume, particle size, and shape, we utilize both static and dynamic tactile sensing. We use different robot actions to interact with the container to estimate the properties. 
We mount a F/T sensor at the robot's wrist to get precise force and torque measurements. We attach a GelSight tactile sensor \cite{dong2017improved} and a newly designed high-speed GelSight tactile sensor on a two-fingered gripper to collect the fingertip tactile signals. The grasping force is set to be $25$ N to prevent slippage between the bottle and the fingertip. The specific setup will be introduced in Sec. \ref{sec: setup}.

\begin{figure}[t]
    \centering
    \includegraphics[width = 0.99\linewidth]{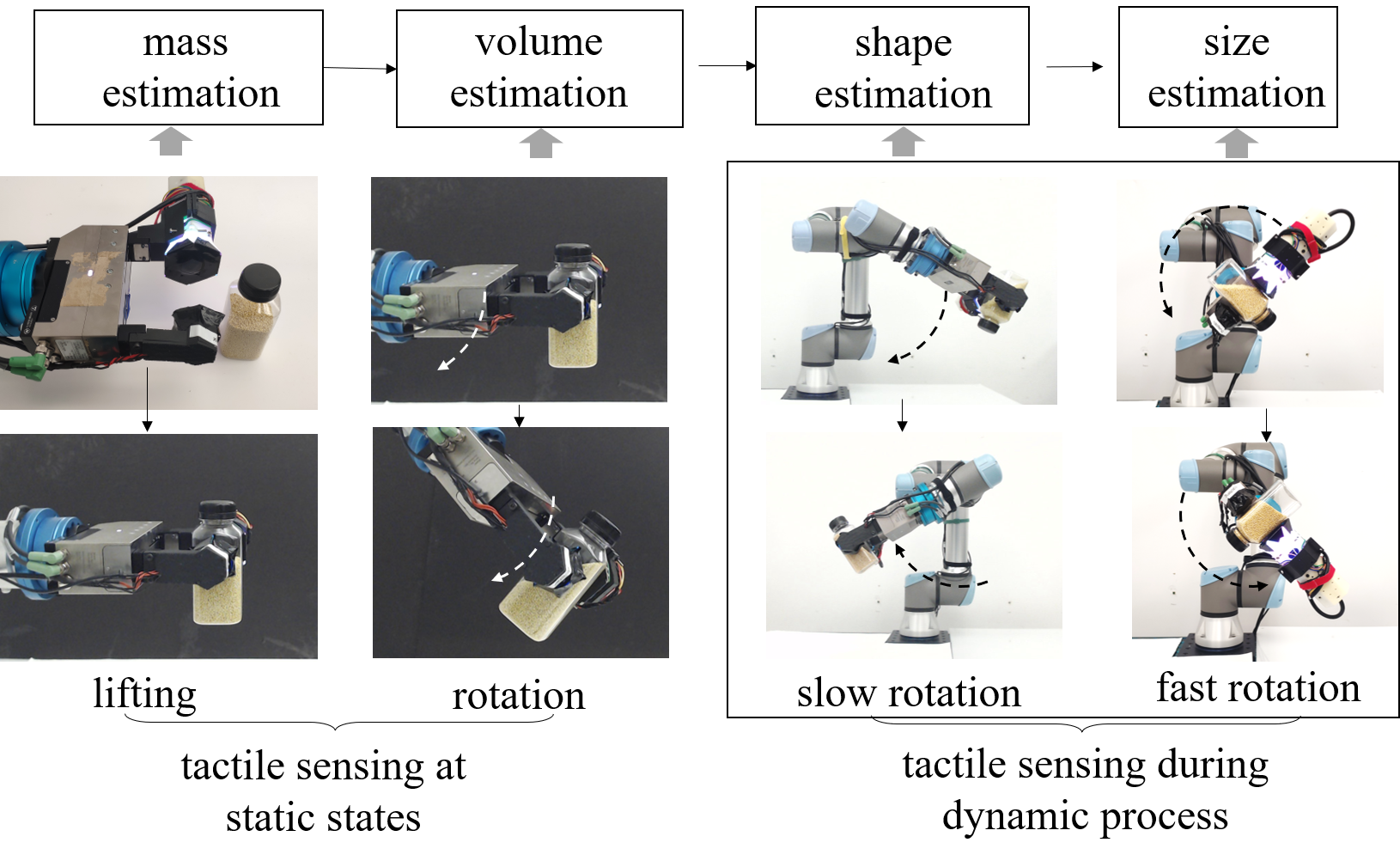} 
    \caption{The pipeline of particle property estimation. We take four actions to activate the container-particle system to estimate different properties. Specifically, the content mass and height are estimated from the tactile measurements at static states, while the particle size and shape are estimated from the tactile measurements during the dynamic process.}
    \label{fig:pipeline}
\end{figure}

\subsection{Mass Estimation and Volume Estimation}
\label{mass_volume}

We estimate the content $M_p$ and the content volume $V_p$ using the wrist-mounted F/T sensor.

To estimate $M_p$, we measure the force change in the vertical direction $\Delta F_z$ after lifting the container. Then 
\begin{equation}\label{eq:mass}
M_p = \frac{\Delta F_z}{g} - M_c
\end{equation}
where $M_c$ is container mass, which is measured in advance.

\begin{figure}
    \centering
    \includegraphics[width = 0.80\linewidth]{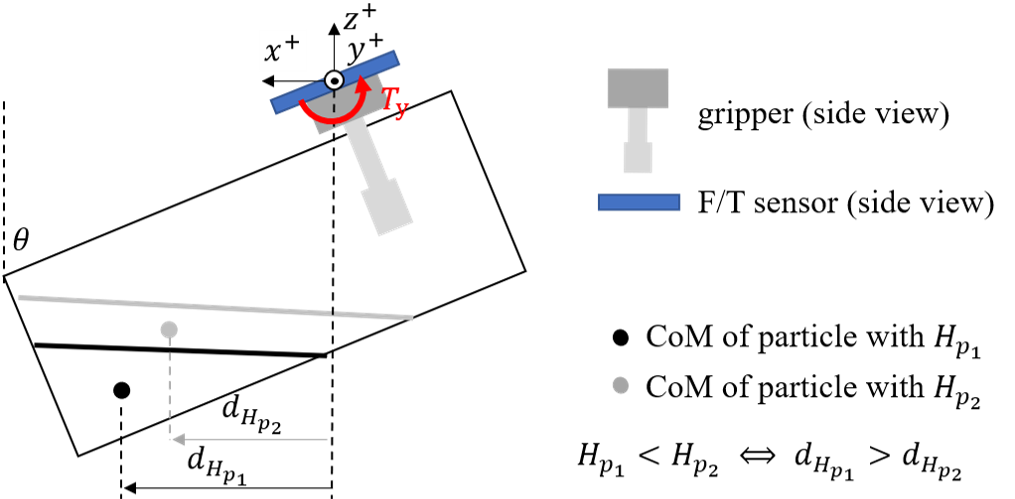}
    \caption{At a fixed tilting angle $\theta$, the measured torque $T_y$ is determined by content mass $M_p$ and content height $H_p$. \wenzhen{Caption is not clear}}
    \label{fig:volumepipeline}
\end{figure}

We estimate content height $H_p$ to represent $V_p$. When tilting the container, particles with different $H_p$ have different center of mass (CoM) locations, resulting in different wrist torque. Specifically, after lifting the container vertically, the robot rotates the container with $\theta$ and shakes the container to make the particle stacking surface horizontal, as shown in the second column in Fig. \ref{fig:pipeline}. After rotation with $\theta$, the measured torque $T_y$ is determined by the content mass $M_p$ and the particles CoM x-position $d_{H_p}$. Additionally, $d_{H_p}$ is monotonically correlated to content height $H_p$, as shown in Fig. \ref{fig:volumepipeline}. Therefore, we have $H_p = H_p(T_y, M_p, \theta)$. $H_p(T_y, M_p, \theta)$ is non-linear due to the irregular wedge shape of the particle pile, as discussed in \cite{matl2019haptic}. Here we utilize the Multi-Layer Perception (MLP) to learn this function.


In our method, we ask the robot to rotate the container to $30$, $45$, and $60$ degrees and measure the torque $T_y$ before shaking and after shaking at each angle. Those measured torques and the estimated particle mass $\Tilde{M_p}$ are used as the features to input to a 4-layer MLP to estimate $H_p$ and $V_p$. The size of two hidden layers of the MLP is set to be $16$ and $4$. We use MSE loss and Adam optimizer to train it.

\subsection{Size Estimation and Shape Estimation} 
\label{section: shape_size}
Then we estimate $D_p$ and $S_p$ by the newly designed high-speed GelSight, a vision-based tactile sensor, which will be introduced in detail in Sec. \ref{sec: design}. This sensor mainly consists of a soft elastomer with printed markers and an embedded camera. After making contact with objects, the soft elastomer deforms and the markers move, both of which will be captured by the embedded camera. Typically, the marker's motion in the images shows the local shear force applied to the soft elastomer. Additionally, we utilize physical inspirations and experiment findings to relate particle size and shape to two macro-scale particle behaviors: vibration intensity during colliding and stacking pattern during rotation. Then we design two robot motions and utilize dynamic tactile sensing to extract vibration-related features and topple-related features to estimate particle size and shape.



\begin{figure}
    \centering
    \includegraphics[width = 0.98\linewidth]{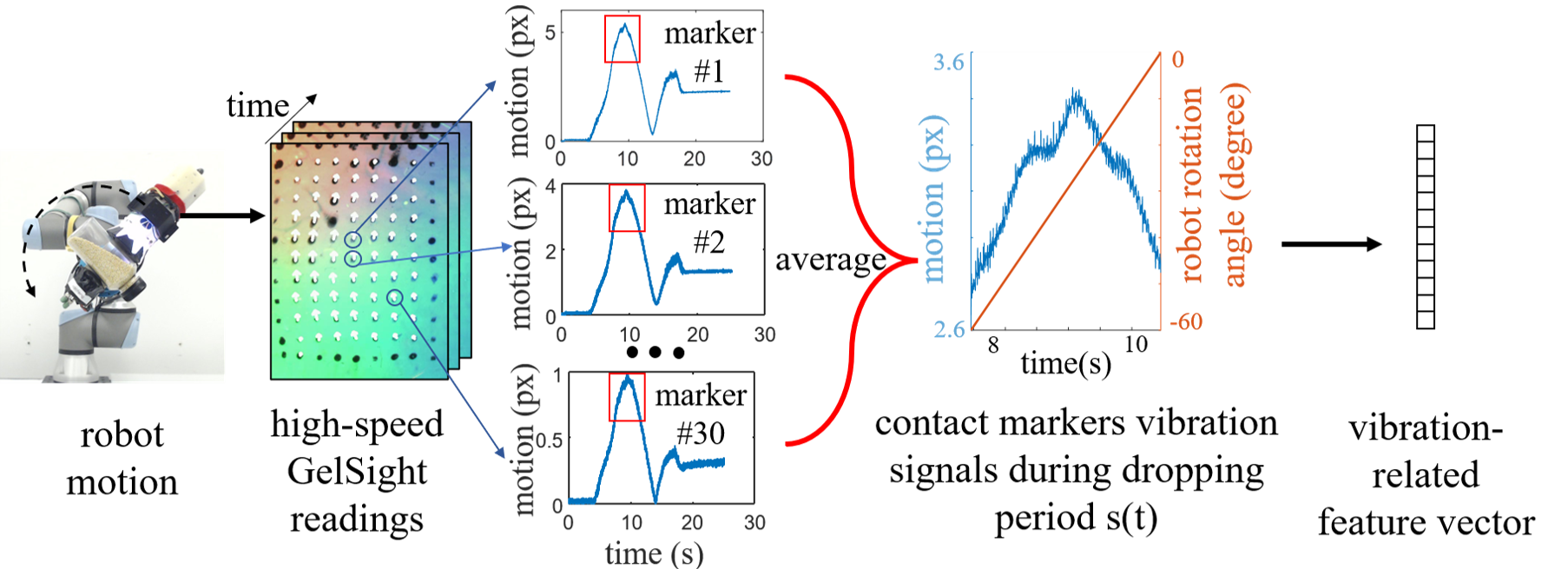}
    \caption{The pipeline of extracting the vibration-related features. The robot rotates the container to make particles contact with the container wall and generates vibration. We collected the vibration signals by a custom-designed high-speed GelSight. We take the average of the contact markers' motion during the dropping period as the vibration signals and extract vibration-related features from it.\wenzhen{Text in this figure is too small. It's hard to read the figure}\feng{how about now?}\wenzhen{better to be larger}}
    \label{fig:vibration_intensity_pipeline}
\end{figure}

\begin{figure}
    \centering
    \includegraphics[width = 0.98\linewidth]{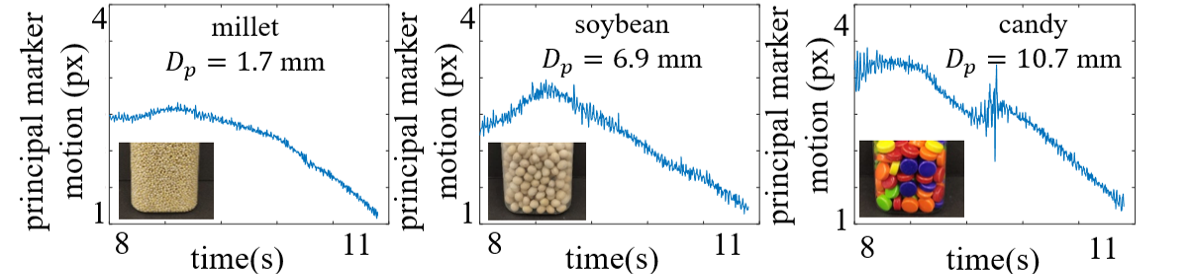}
    \caption{Extracted vibration signals $s(t)$ of particles of different sizes. Larger particles generate more intense vibrations.\wenzhen{text too small}}
    \label{fig:vibration_intensity}
\end{figure}

The first robot motion is to rotate the container from a nearly upside-down pose to an upward pose, shown as the fast rotation motion in Fig. \ref{fig:pipeline}. It aims to initiate collisions between particles and the container wall to generate vibration signals. Intuitively, the intensity of this vibration signal is positively correlated with particle size: large particles will lose more kinetic energy after colliding with the container walls, leading to a stronger contact force, and resulting in higher vibration intensity. Specifically, we rotate the container's long axis from $-135$ to $135$ degrees relative to the upward direction with $15$ degree/s. We collect the high-speed GelSight images during the period when the container's long axis is $-60$ to $0$ degrees relative to the upward direction. This is the duration that the particles fall and collide with the bottom of the container. We extract all the markers' position sequences from the high-speed GelSight images and use the top $30$ markers with the largest motion as the markers inside the contact area. We average their motions as the principle vibration signal $s(t), t\in[1, T]$. Fig. \ref{fig:vibration_intensity} shows examples of the vibration signal collected with different particles and also shows that larger particles usually have vibrations with higher intensity. We design two vibration-related features to capture the vibration intensity: $ v_{1_a} = \sum_{t=a+1}^T \lvert s(t)-s(t-a)\rvert, a = [1,\dots 50] $ and $v_{2_a} = \sum_{t=a+1}^{T-a} \lvert 2s(t)-s(t-a)-s(t+a)\rvert, a = [1,\dots 50]$. Intuitively, these features quantify the local variability of $s(t)$ over a specific time scale defined by the step size $a$. The vibration-related feature extraction pipeline is shown in Fig. \ref{fig:vibration_intensity_pipeline}.
\wenzhen{these are too details about the method. You need to give an overview of the major algorithm of your method, explain why it works, and then go to the details} 

\begin{figure}
    \centering
    \includegraphics[width = 0.8\linewidth]{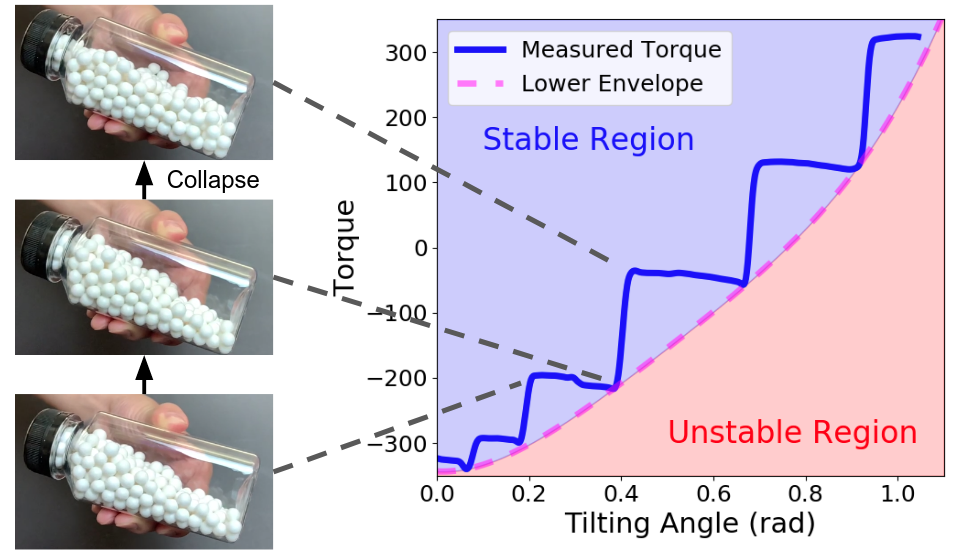}
    \caption{Step-like torque signal caused by the collapse. When the tilting angle enters the unstable region, the particle stack will collapse to a new stacking state, and the process will make a step-like shape in the recorded torque signal. 
    The lower envelope of the step-like signal is the boundary between the stable and unstable region.}
    \label{fig:stack_pattern}
\end{figure}

\begin{figure}
    \centering
    \includegraphics[width = 0.98\linewidth]{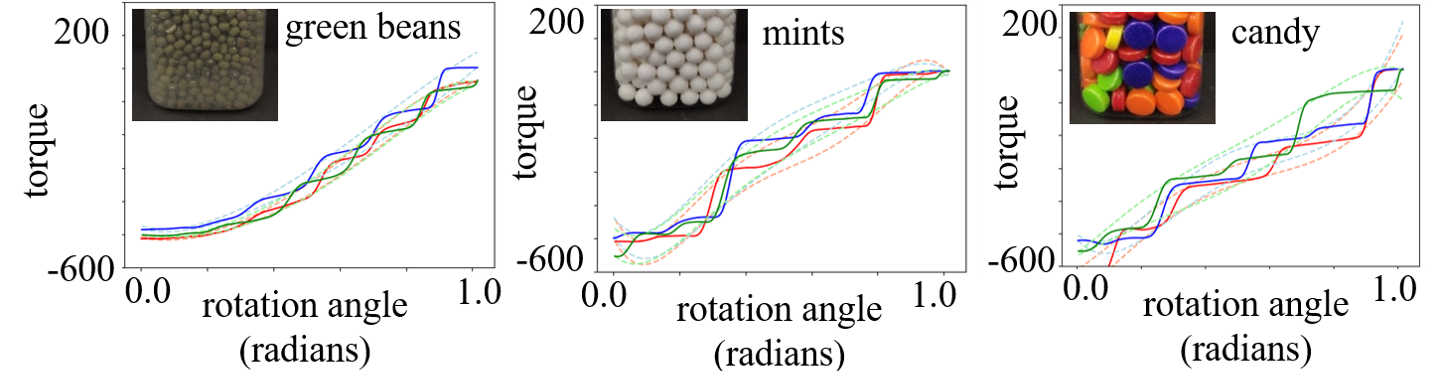}
    \caption{The step-like torque signal when rotating the container with different common particles. The signals with different colors in each subfigure show the repeated trial data of the same particles with the same content volume and the dashed lines are the fitted envelopes of those data. The exact step-like torque signal differs between repeated trials but shares similar envelopes.}
    \label{fig:collapse_pattern}
\end{figure}

The second robot motion is to slowly tilt the container to explore the stability of the particle stack at different tilting angles, shown as the slow rotation motion of Fig. \ref{fig:pipeline}. During rotation, after the tilting angle exceeds a certain threshold (the angle of repose (AoR) introduced in Sec. \ref{sec: relatedworks}), the particle stack collapses and reaches a new stacking state. The process repeats. Different particle stacking state has different particle CoM, causing different fingertip torque. Therefore, the discontinuous particle stack collapse causes the plot of the fingertip torque to the tilting angle to have a step-like shape, as shown in Fig. \ref{fig:stack_pattern}. The lower envelope of the torque signal (dotted pink line in Fig. \ref{fig:stack_pattern}) describes the maximum tilting angle for each stacking state that maintains stability. In practice, the exact step-like torque signal differs between repeated trials since the initial particles' states can be slightly different; but its lower envelope stays similar across trials (Fig. \ref{fig:collapse_pattern}) since it reflects particle AoR that is unaffected by the initial particles' state. Similarly, the upper envelope reflects the state after the particles collapse. Intuitively, both envelopes are related to the particle AoR and therefore highly relate to the particle shape and size. For example, spherical particles collapse easier and formulate the upper and lower envelope with less difference, as shown in Fig. \ref{fig:collapse_pattern}.

\wenzhen{the description is not clear enough}

Specifically, the robot rotates the container from $-60$ to $60$ degrees (third column of Fig. \ref{fig:pipeline}) and records the high-speed GelSight images during this process. We extract the rotation patterns of the high-speed GelSight markers to estimate the fingertip torque (note that the unit of measured torque is sensor-related and is not ${N}\cdot m$). We then extract the lower and upper envelope of the step-like torque signal and equally sample $100$ points from both envelopes to represent the topple-related feature.

Both the two features: vibration-related features and topple-related features are correlated with the particle properties. Here we utilize learning methods to model the complex relationship between the features and the particle size and shape. Specifically, we stack the vibration-related features ($100$ dimensions), and the topple-related features ($200$ dimensions), together with the estimated content mass and content volume from Sec. \ref{mass_volume} as the total feature vectors. We input those features into a 4-layer MLP to estimate $D_p$ and into the Random Forests classifier for shape classification. The size of two hidden layers of the MLP is set to be $16$ and $4$. We use MSE loss and Adam optimizer in Pytorch to train it.

\section{HIGH-SPEED GELSIGHT TACTILE SENSOR}
\label{sec: design}



In our method described in Sec. \ref{sec: method}, we relate particle size to fingertip vibration intensity. To sense the fingertip vibration signals, we design and fabricate a new tactile sensor: the high-speed GelSight tactile sensor. GelSight is a commonly used vision-based tactile sensor \cite{dong2017improved}. In their design, the sensor captures tactile images with $640\times480$ pixels resolution at $30$ Hz. However, the vibrations signal has a higher frequency and the human finger can sense signals up to approximately $700$ Hz \cite{dahiya2009tactile}. In our experiment, we found the frequency of the vibration signals caused by particle contact is over $100$ Hz. Based on that, we request a sensor that has a higher sampling rate.

In our design, we keep the basic structure of GelSight, but use a high-speed camera (A5031CU815 from HUARAY Tech.) to capture the high-frequency tactile signals. The camera captures images as fast as $815$ Hz. In this work, we capture the images at $800$ Hz with $640\times480$ spatial resolution. Additionally, we attach a $12$ mm lens to the camera and its field of view is about $18 \text{ mm}\times14 \text{ mm}$. 3 RGB LEDs of around $100$ lm are used to the lighting the gel pad. With respect to the gel pad, we fabricate a dome-shaped silicone pad for robust grasping and transfer a marker array with $1.7$ mm spacing to the surface of the gel pad. In total, there are around 70 markers in the field of view for tracking to represent the local motion of the gel pad. Fig. \ref{fig:experimentsetup}b shows the mechanical designs and real images of our tactile sensor. The massive tactile data will bring a higher computational burden when processed online, so we take all processing offline. Experiments in Sec. \ref{sec: comparison} show the benefits of using our newly designed sensor in this task.




\section{EXPERIMENT RESULTS}
\label{sec: results}

We conduct experiments to evaluate our pipeline's performance in estimating the four properties of the solid particles inside the container. In this section, we show the results for both seen and unseen particles. We also evaluate the influence of three key components of our system: the choice of sensors, actions, and features to extract from the raw data. 



\subsection{Experiment Setup and Dataset}
\label{sec: setup}

The robot experiment setup is shown in Fig. \ref{fig:experimentsetup}a. We use a 6 DoF robotic arm (UR5E from Universal Robotics) and a parallel gripper (WSG50 from Weiss Robotics) to grasp the bottle. One Fingertip GelSight \cite{dong2017improved} and one newly designed High-Speed GelSight are mounted on the gripper. A 6-axis Force/Torque sensor (NRS-D50 from Nordbo Robotics) is mounted at the wrist.

\begin{figure*}
    \centering
    \includegraphics[width = 0.9\linewidth]{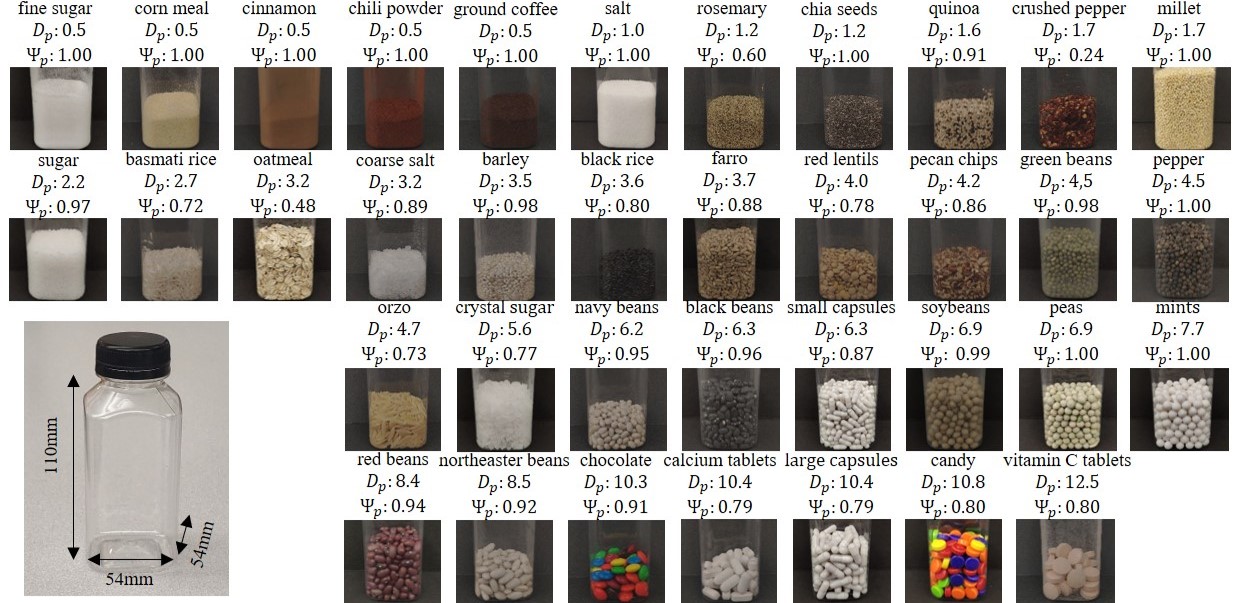}
    \caption{The container and the $37$ common particles in our dataset. The particles are selected to represent a wide variety of common particles.}
    \label{fig:dataset}
\end{figure*}


\begin{figure}
    \centering
    \includegraphics[width = 0.99\linewidth]{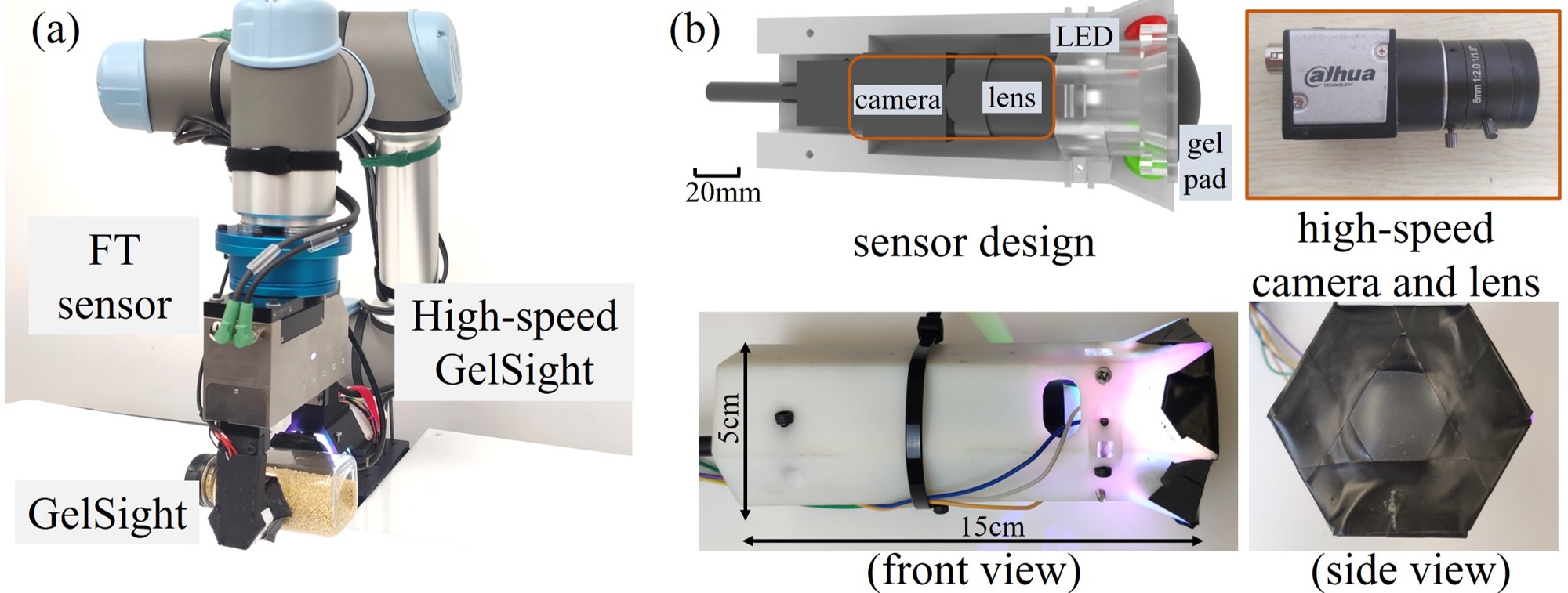}
    \caption{(a) A 2-fingered gripper with one GelSight and our designed high-speed GelSight is used to grasp the bottle. A F/T sensor is installed to precisely measure the wrist force. (b) We design a high-speed GelSight tactile sensor to sense the vibration tactile signals.}
    \label{fig:experimentsetup}
\end{figure}


We collect a dataset of $37$ different common solid particles, as shown in Fig. \ref{fig:dataset}. These particles are different in terms of density, size, and shape. They are selected to cover a wide variety of particles in daily life. The particles are stored inside a common plastic cuboid bottle. In this work, we only focus on this particular type of container. We collect data at three levels of content heights $H_p$: $30 \text{mm}, 50\text{mm}, \text{and } 70\text{mm}$.
We perform the data collection process three times for each particle-height setting, formulating a dataset with $333$ data points. 


\subsection{Property Estimation with Seen Particles}

\begin{figure}
    \centering
    \includegraphics[width = 0.95\linewidth]{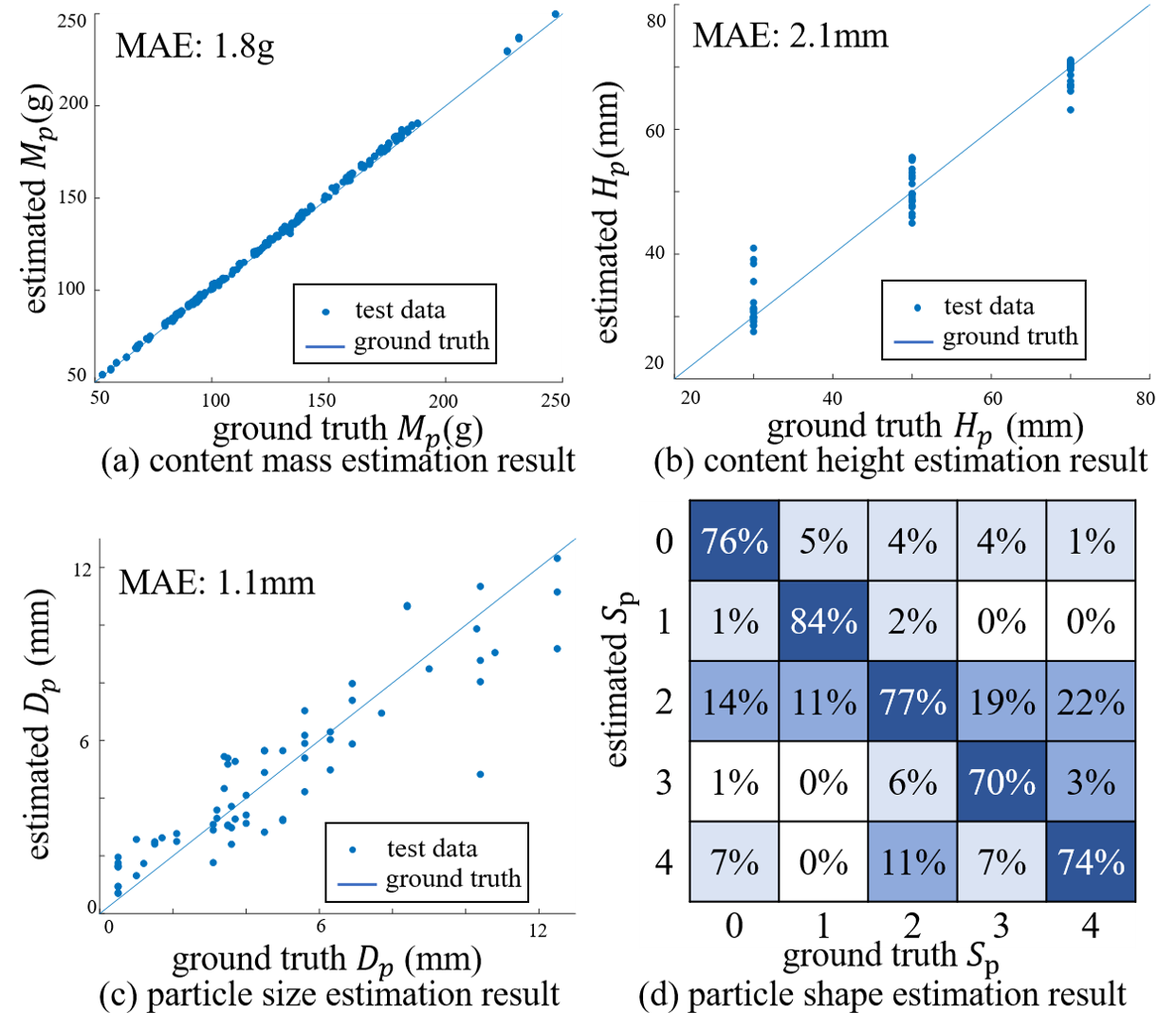}
    \caption{Particle properties estimation results with seen particles.}
    \label{fig:property_estimation_result}
\end{figure}

We first evaluate our method for estimating particle properties on seen particles. The experiment is conducted on all the $37$ particles in the dataset. Here we randomly split the $333$ data points into $80\%$ training data points and $20\%$ testing data points. The results are shown in Fig. \ref{fig:property_estimation_result}.

\subsubsection{Mass Estimation}
Our method estimates $M_p$ with mean absolute error (MAE) of $1.8$g, and mean absolute percentage error (MAPE) of $1.4$\%.

\subsubsection{Volume Estimation}
We estimate $V_p$ by estimating $H_p$. Our approach estimates $H_p$ with MAE of $2.1$mm and MAPE of $4.3$\%, which represents the $V_p$ estimation MAE is about $6.1$ml and MAPE is $4.3$\%. Note that the ground truth of $H_p$ has a measurement error of roughly $1$ mm.

\subsubsection{Size Estimation}
The particle size $D_p$ estimation MAE is about $1.1$ mm, which is around the size difference between barley and green beans. For most spherical particles, the estimation error is small, while for the irregular particle, the estimation error is larger. There is a measurement error of $0.5$ mm for the groundtruth measurement. 
The complex particle dynamic increases the particle behavior variance under small disturbances. In addition, multiple other particle parameters such as texture or friction coefficient would also influence the particle behavior. Some of the daily particles such as rosemary or crushed pepper are very light, decreasing the sensory data magnitude. All the above contribute to the estimation error.

\subsubsection{Shape Estimation}
We classify the particles into different shape groups, as defined in Sec. \ref{sec: problemstatement}. The classification accuracy is $75.6$\%. Our method recognizes most of the particles well. Most of the misclassification of particles happens to the particles whose shape is closer to spherical. This is because the sphericity difference of those particles is smaller and the other particle property such as stickiness would also affect the particle behavior.


\subsection{Property Estimation with Unseen Particles}
To evaluate the generalizability of our approach, we utilize it to estimate the properties of unseen particles. Among our dataset, we select $5$ types of particles for testing: ground coffee, rosemary, orzo, navy beans, and green beans. We use the data collected from the other $32$ particle types for training. To test the generalizability with volume estimation, for each test particle, we also collected 6 data points with different random volumes. So we have $288$ training data points and $75$ test data points.

\begin{figure}
    \centering
    \includegraphics[width = 0.95\linewidth]{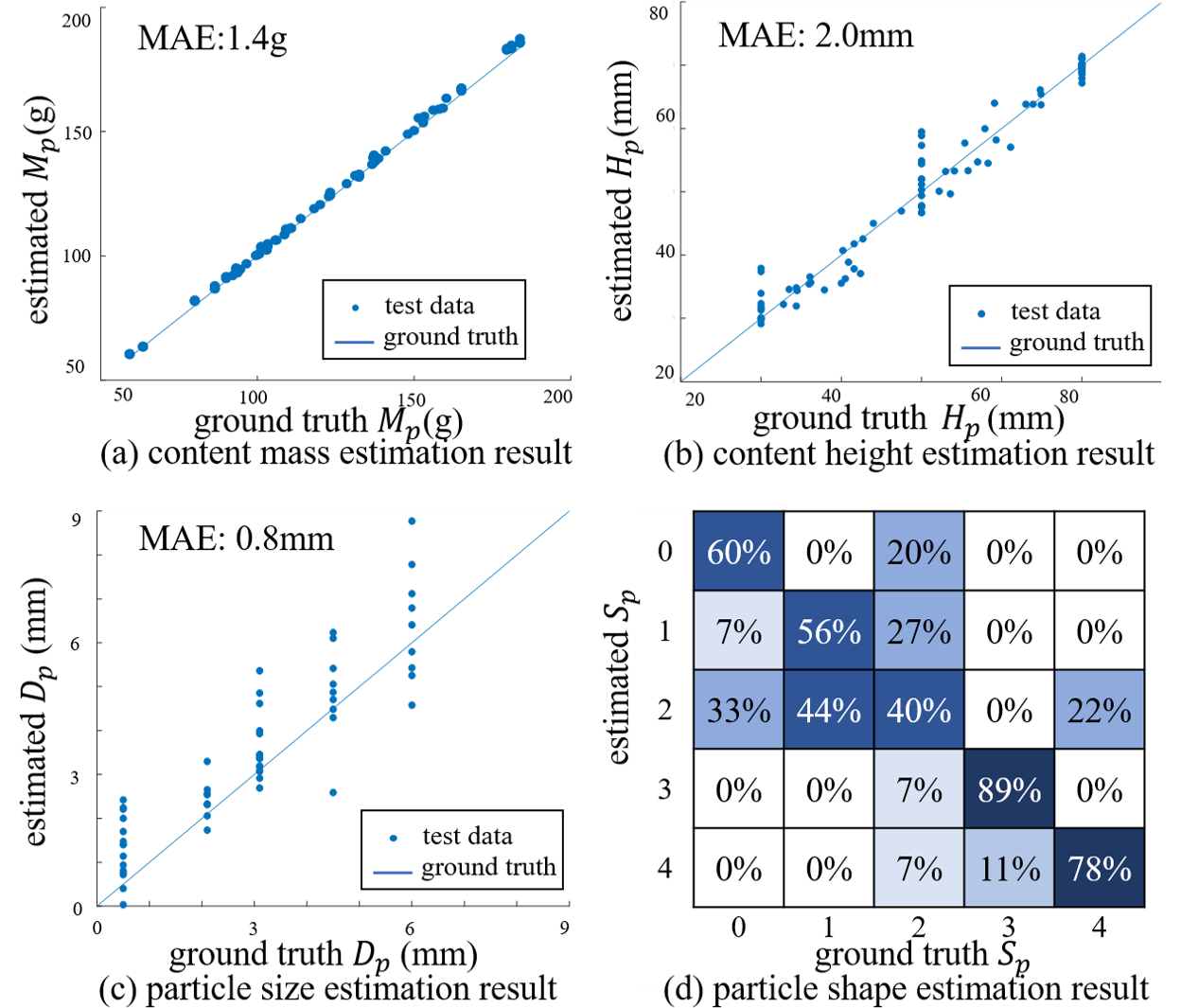}
    \caption{Particle properties estimation results of unseen particles with unseen volume.}
    \label{fig:generalization_result}
\end{figure}

The results are shown in Fig. \ref{fig:generalization_result}: MAEs of $M_p$, $H_p$, and $D_p$ estimation are $1.4$g, $2.0$mm, and $0.8$ mm, respectively. The particle shape classification accuracy is $64$\%. The result shows that our approach can estimate unseen particles' properties. For most particles, the estimation error of mass, volume, and size is in a similar range to the error testing on seen particles. The algorithm works well to recognize the shape of spherical particles such as navy beans and green beans but performs worse when estimating the shape of irregular particles. Most of the misclassification happens to adjacent groups, which are harder to distinguish.  


\begin{table*}[]
    \caption{Comparison of using different sensors, different actions, and different signal processing methods}
    \label{tab:comparison}
    \centering
    \begin{tabular}{c|c|c|c|c}
    \hline
    Tactile sensor & Actions & Signal processing method & size estimation MAE (mm) & shape classification accuracy \\
    \hline \hline
    Normal GelSight & \makecell{two dynamic robot motion \\ (our action)} & \makecell{feature engineering \\ (our method)} & 1.5 & 63.3\%\\
    \hline
    High-speed GelSight & horizontal shaking \cite{chen2016learning} & spectrum features \cite{chen2016learning} & 2.3 & 36.1\% \\
    \hline 
    High-speed GelSight & \makecell{two dynamic robot motion \\ (our action)} & MFCC + LSTM \cite{jonetzko2020multimodal} & 2.9 & 37.3\% \\
    \hline
    \textbf{High-speed GelSight} & \makecell{ \textbf{two dynamic robot motion} \\ \textbf{(our action)}} & \makecell{ \textbf{feature engineering} \\ \textbf{(our method)}} & \textbf{1.1} & \textbf{75.6\%} \\
    \hline
    \end{tabular}

\end{table*}

\subsection{Comparison with Other Methods}
\label{sec: comparison}
The good performance of our method relies on three key components: the use of high-speed GelSight, the optimized actions, and the physical-inspired feature extraction process. Here we evaluate the necessity of those components by comparing the property estimation results using different settings, and the results are shown in Table \ref{tab:comparison}. 

\noindent\textbf{Choice of Sensor: }
We first make a comparison with if we only use the traditional GelSight sensor~\cite{dong2017improved} with a sampling rate of $30$ Hz. Using this setting, we lose some high-frequency vibration features during the fast rotation action. The result shows that using high-speed GelSight improves both size estimation and shape estimation accuracy. It also shows the high-frequency vibration features benefit the task.

\noindent\textbf{Choice of Action: }
In our method, we designed two rotation motions based on the desired perception of particle macro-scale behavior to estimate particle size and shape. Here we compare it with the baseline motion: shaking in one direction, which is a common action used to classify solid particles in previous work \cite{chen2016learning}, \cite{eppe2018deep}. Specifically, we ask the robot to shake the container horizontally at $2$ Hz and record dynamic tactile signals by high-speed GelSight. The magnitude of each frequency band is commonly used as the spectrum features of the vibration signal \cite{chen2016learning}. Here we extract the spectrum features from tactile signals of horizontal shaking and input it to a 4-layer MLP to estimate particle size and shape. The results show that by using our actions, the property estimation results improved significantly.

\noindent\textbf{Choice of Features: }
In our method, based on the physical inspiration, we choose vibration-related features and topple-related features for estimating the particle shape and size. We manually designed the signal processing and feature extraction process. 
Instead, a common way to process temporal data is to use recurrent neural networks. Specifically, previous works \cite{eppe2018deep}, \cite{jonetzko2020multimodal} on particle classification employed Mel Frequency Cepstral Coefficients (MFCC) preprocessing to the signal and then input the Mel coefficients to LSTM for classification. Here we compare our results with their method. We apply MFCC to the tactile signals we collected during the two exploration processes mentioned in Sec. \ref{section: shape_size} and use LSTM for the estimation of particle size and shape.
It turns out that the estimation error is significantly larger than ours. This shows that the physics-inspired features we use effectively extracted the important information relevant to the particle dynamics from the tactile signal.

\subsection{Sugar Humidity Estimation}

Other than the properties we described before, some properties are specific to only certain types of particles but are highly useful at the function level. A typical example is the humidity of sugar, which helps humans to handle them properly. 
In this subsection, we apply our method to estimate sugar humidity. According to \cite{fraysse1999humidity}, the granular media AoR is also related to particle humidity, inspiring us to use the topple-related features to infer the sugar humidity since these features reflect the particle AoR. Intuitively, higher humidity force sugar to be sticky and hard to collapse, while the dry sugar would flow more like the liquid.
To magnify the collapse pattern of the sticky sugar, we use a rotation motion from $-135$ to $135$ degrees for the exploratory procedure and record the tactile signals by the high-speed GelSight. Then we extract the topple-related features in the same way described in Sec. \ref{section: shape_size}. We then input those features into a 3-layer MLP to estimate the sugar humidity.

\begin{figure}
    \centering
    \includegraphics[width = 0.99\linewidth]{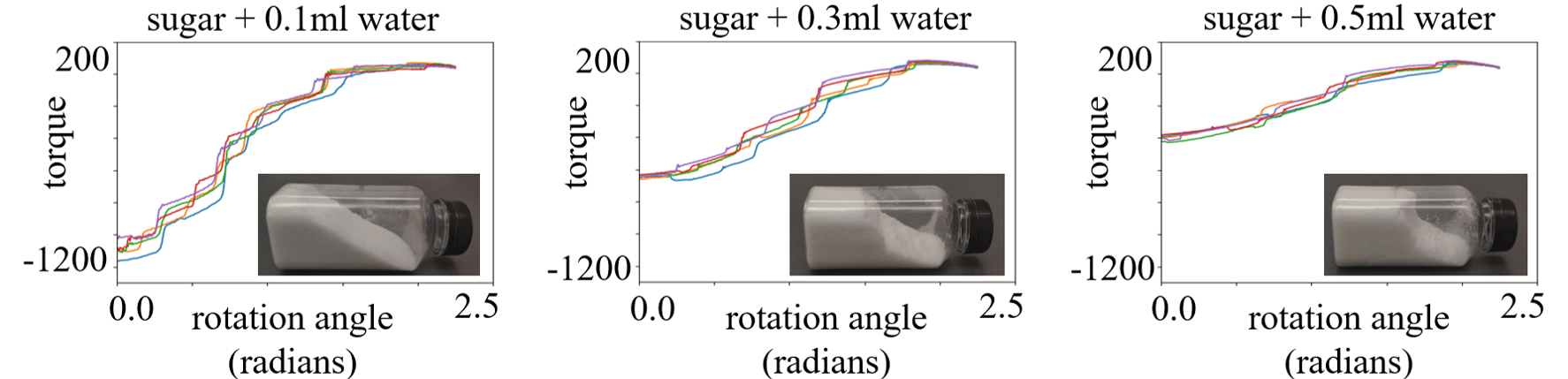}
    \caption{The step-like torque signal when rotating the container with sugar in different humidity. Sugar with higher humidity collapses less because more sugar is stuck to the container wall, leading to less change in the step-like signal. \wenzhen{The caption is unclear}\wenzhen{Text in the figures too small to read}}
    \label{fig:sugar_stacking_pattern}
\end{figure}

\begin{figure}
    \centering
    \includegraphics[width = 0.95\linewidth]{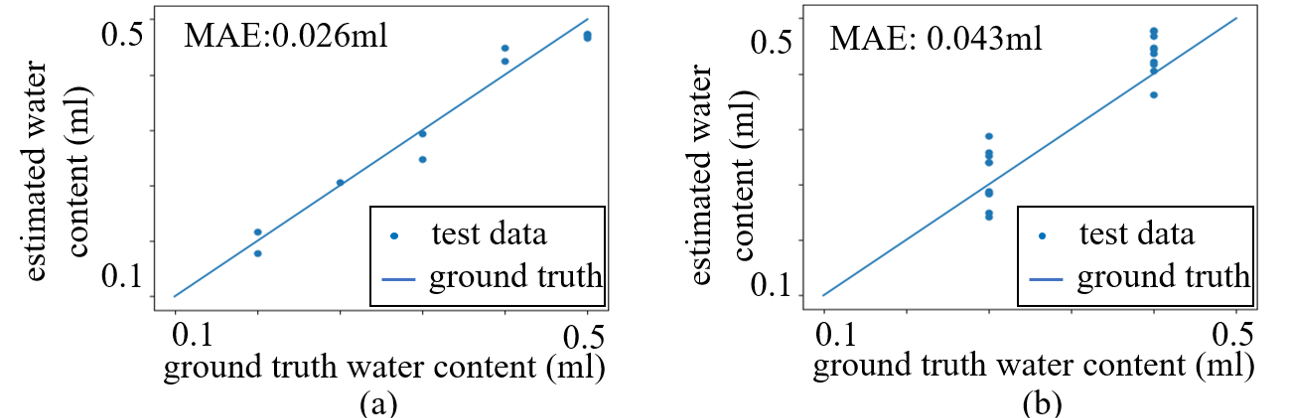}
    \caption{Sugar humidity estimation results. We estimate the humidity by predicting the volume of added water and test on (a) seen humidity and (b) unseen humidity.}
    \label{fig:sugar_result}
\end{figure}

To collect a dataset for the task, we manually drop $0.1$, $0.2$, $0.3$, $0.4$, and $0.5$ ml of water into the $150$ g fine sugar powder to create the sugar with $5$ different humidity levels. We use the volume of water to denote the humidity level in this specific setting. 
Fig. \ref{fig:sugar_stacking_pattern} shows the step-like torque signals collected in $3$ groups of humid sugar. Sugar with higher humidity has a larger portion to be sticky to the container wall and never collapses, resulting in a lesser variation in the step-like signals. We repeat the exploratory procedure on the $5$ groups of humid sugar with 10 trials. Fig. \ref{fig:sugar_result}a shows the estimation result if we randomly split the whole dataset into $80$\% training dataset and $20$\% testing dataset. Our estimation of the added water has an MAE of $0.026$ ml. Fig. \ref{fig:sugar_result}b shows the estimation result if we train on humid sugar with $0.1$, $0.3$, $0.5$ ml of water and test on humid sugar with $0.2$, $0.4$ ml of water. Our estimation of the added water has an MAE of $0.043$ ml. This experiment shows the potential of our methodology to estimate other particle properties.

\feng{include the insight in the captions as well}
\feng{modified}

\wenzhen{caption of fig.14 is unclear}
\feng{modified}



\section{CONCLUSION}
\label{sec: conclusion}

In this work, we present an approach to estimate the properties of solid particles in the container. We combine static and dynamic tactile sensing to estimate content mass, content volume, particle size, and particle shape. We design a sequence of exploratory procedures to interact with the container to estimate those four properties. In the experiment, our approach can estimate those properties, achieving an accuracy of $1.8$ g, $6.1$ ml, $1.1$ mm, and $75.6$\% for mass, volume, size, and shape estimation. Additionally, we show that our method can generalize to unseen daily particles with unseen volumes. We also show that the physical-inspired actions and features can be used to estimate sugar humidity. We believe our explorations on solid particle property estimation will help with different manipulation tasks in daily life and industry.




\bibliographystyle{IEEEtran}
\bibliography{root}

\end{document}